\documentclass[acmsmall]{acmart}
\usepackage{xcolor}
\usepackage{graphicx}
\usepackage{array}
\def\rbar#1{%%
  #1 & {\color{red}\rule{#1cm}{8pt}}}
\def\bbar#1{%%
  #1 & {\color{blue}\rule{#1cm}{8pt}}}

\AtBeginDocument{%
  \providecommand\BibTeX{{%
    \normalfont B\kern-0.5em{\scshape i\kern-0.25em b}\kern-0.8em\TeX}}}

\setcopyright{acmcopyright}
\copyrightyear{2020}
\acmYear{2020}
\acmConference[arXiv]{arXiv}{October 2020}{Ithaca, NY}
\acmMonth{10}

\begin{document}

%%
%% The "title" command has an optional parameter,
%% allowing the author to define a "short title" to be used in page headers.
\title[Representation and Storytelling in Me Too]{Demographic Representation and Collective Storytelling in the Me Too Twitter Hashtag Activism Movement}

%%
%% The "author" command and its associated commands are used to define
%% the authors and their affiliations.
%% Of note is the shared affiliation of the first two authors, and the
%% "authornote" and "authornotemark" commands
%% used to denote shared contribution to the research.

%\author{Anonymous CSCW Submission}

\author{Aaron Mueller}
\email{amueller@jhu.edu}
%\orcid{1234-5678-9012}
\affiliation{%
  \institution{Johns Hopkins University}
  \streetaddress{3400 N. Charles St.}
  \city{Baltimore}
  \state{Maryland}
  \postcode{21218}
}

\author{Zach Wood-Doughty}
\email{zach@cs.jhu.edu}
\affiliation{%
  \institution{Johns Hopkins University}
  \streetaddress{3400 N. Charles St.}
  \city{Baltimore}
  \state{Maryland}
  \postcode{21218}}

\author{Silvio Amir}
\email{samir@jhu.edu}
\affiliation{%
  \institution{Johns Hopkins University}
  \city{Baltimore}
  \state{Maryland}
  \postcode{21218}
}

\author{Mark Dredze}
\email{mdredze@cs.jhu.edu}
\affiliation{%
  \institution{Johns Hopkins University}
  \streetaddress{3400 N. Charles St.}
  \city{Baltimore}
  \state{Maryland}
  \postcode{21218}}

\author{Alicia L. Nobles}
\email{alnobles@ucsd.edu}
\affiliation{%
 \institution{University of California San Diego}
 \streetaddress{9500 Gilman Dr.}
 \city{La Jolla}
 \state{California}
 \postcode{92093}}

%%
%% By default, the full list of authors will be used in the page
%% headers. Often, this list is too long, and will overlap
%% other information printed in the page headers. This command allows
%% the author to define a more concise list
%% of authors' names for this purpose.
\renewcommand{\shortauthors}{Mueller et al.}

%%
%% The abstract is a short summary of the work to be presented in the
%% article.
\begin{abstract}
  The \#MeToo movement on Twitter has drawn attention to the pervasive nature of sexual harassment and violence. While \#MeToo has been praised for providing support for self-disclosures of harassment or violence and shifting societal response, it has also been criticized for exemplifying how women of color have been discounted for their historical contributions to and excluded from feminist movements. Through an analysis of over 600,000 tweets from over 256,000 unique users, we examine online \#MeToo conversations across gender and racial/ethnic identities and the topics that each demographic emphasized. We found that tweets authored by white women were overrepresented in the movement compared to other demographics, aligning with criticism of unequal representation. We found that intersected identities contributed differing narratives to frame the movement, co-opted the movement to raise visibility in parallel ongoing movements, employed the same hashtags both critically and supportively, and revived and created new hashtags in response to pivotal moments. Notably, tweets authored by black women often expressed emotional support and were critical about differential treatment in the justice system and by police. In comparison, tweets authored by white women and men often highlighted sexual harassment and violence by public figures and weaved in more general political discussions. We discuss the implications of work for digital activism research and design including suggestions to raise visibility by those who were under-represented in this hashtag activism movement. \textbf{Content warning: this article discusses issues of sexual harassment and violence.}  %Finally, people of color engaged in higher rates of retweeting, potentially reflective of more community-based dialogue. 
\end{abstract}

%%
%% The code below is generated by the tool at http://dl.acm.org/ccs.cfm.
%% Please copy and paste the code instead of the example below.
%%


%%
%% Keywords. The author(s) should pick words that accurately describe
%% the work being presented. Separate the keywords with commas.

%%
%% This command processes the author and affiliation and title
%% information and builds the first part of the formatted document.
\maketitle

\section{Introduction}
Tarana Burke aimed to break through the silence surrounding sexual harassment and violence by initiating Me Too, an offline grassroots movement, in 2006. This movement focused on increasing resources and building a community of peer advocates for survivors of sexual violence---especially for low-wealth communities of black women---in part by engaging in storytelling and data sharing about sexual violence \cite{taranaburke,metooorigins}. Alyssa Milano elevated the visibility of the movement when she tweeted a request for solidarity of victims by asking people to simply reply with \textit{``\#MeToo''} \cite{metooorigins} on October 15, 2017. This request sparked a viral hashtag activism movement across multiple social media platforms with victims responding in solidarity, many recounting personal experiences. This resulted in over half a million \#MeToo tweets within the first 24 hours after Milano's tweet \cite{otherhashtags}.

The online movement has been praised for opening the door to the long-held silence among victims of sexual harassment and violence, enabling people from disparate geographic regions and cultures to participate broadly in both collective storytelling online and collective action offline, shifting societal response to sexual harassment and violence. However, the online movement has been rightfully criticized of dismissal of the contributions to feminist movements from women of color, as well as generally under-representing the voices of women of color \cite{intersectionalmetoo2,trott2020networked,leung2019metoointersectional,bowman2020maximizingdiss}---a significant criticism, especially given the original foundation of Me Too by a woman of color and intent of Me Too to help marginalized communities of color \cite{taranaburke,metooorigins}.

While the movement has drawn attention from scholars focusing on a myriad of aspects, including the content and emotions expressed in tweets \cite{manikonda2018twittersparking}, how tweets revealed first-person accounts of sexual violence and received support \cite{Modrek_JMIR_2019,Alaggia_2020}, how tweets communicated victimhood \cite{Simpson_CSCW_2018}, and the role of reciprocal disclosure as a motivator for self-disclosure \cite{Gallagher_CSCW_2019}, less is known about ``{\em who}'' is represented in this online movement. That is, what voices contributed to the hashtag activism movement and what issues were salient to these voices? Attempts to understand representation are currently limited to an examination of types of social support across binary gender during the first six months of the movement \cite{hosterman2018twitter} or recalling the frequency of disclosure of sexual assault or abuse across binary gender, race, and ethnicity during the initial week of the movement \cite{Modrek_JMIR_2019}. 

Examining who participated in the \#MeToo movement is challenging, but understanding representation is critical to effecting change that reaches all individuals, especially those that are most vulnerable and often excluded from discourse \cite{Jackson_hashtag_activism_2020}. Moreover, critical race theory highlights the necessity of marginalized voices, which are best heard through storytelling of their lived experience \cite{Ogbonnaya_CriticalRaceTheory_CHI_2020}. Our study aims to examine large-scale patterns in representation across intersecting identities of gender, race, and ethnicity (i.e., the framework of intersectionality \cite{intersectionality}) which have been critically important in historical movements \cite{intersectionalmetoo1,intersectionalmetoo2,demographicsblm}. We also aim to investigate topics salient to these groups through the following research questions. 

\textbf{RQ1: {\em Who actively tweeted} in the \#MeToo Twitter hashtag activism movement?} We automatically inferred perceived demographic identities including gender, race, and ethnicity, and the intersection of these, for user accounts tweeting about \#MeToo and related topics (e.g., sexual assault). Note that we are inferring perceived demographics, which do not necessarily align with self-identified demographics. We discuss these limitations in the paper.

\textbf{RQ2: {\em What stories were shared} and how did they vary temporally and across inferred demographic identities?} We conducted a topic model analysis to examine the stories and comments shared in \#MeToo tweets, sharing edited tweets to contextualize these automatically derived topics. We contextualize these stories by finding the most probable and most distinctive topics per-demographic. To analyze these topics of conversation across time, we also analyze which topics are most common at specific pivotal points during the movement.

\textbf{RQ3: {\em What hashtags were associated} with this movement and how did they vary temporally and across inferred demographic identities?} We perform a similar analysis to the previous topic analysis; namely, we observe which hashtags were most common overall in \#MeToo and related tweets, as well as which hashtags were most common across demographic groups and at pivotal points in the movement.

%\textbf{RQ4: {\em How did active participation vary} across inferred demographic identities?} For this analysis, we observe what types of tweets were most common among demographic groups. For example, do women tend to share their own ideas and stories more than men by publishing original tweets more frequently? Do women and/or people of color share others' stories more often than men and/or white users---for example, by retweeting and replying more often?

\subsection{Contributions}
Our work extends on previous studies that examined how collaborative communication technologies were leveraged for digital activism to examine how intersectional identities leveraged Twitter to contribute to the framing of and participate in the \#MeToo hashtag activism movement. More specifically, rather than \textit{a priori} selecting a particular construct to examine, we examined large-scale patterns in the topics discussed and hashtags used by the intersected identities of gender, race, and ethnicity during the first year of \#MeToo---a first in examining contributions from intersected identities for this movement. We discuss how these identities contributed different narratives to frame the movement (e.g., critical narratives about differential treatment in the justice system and by police), co-opted the movement to raise visibility in parallel ongoing movements (e.g., \#Brutality, \#ClimateChange), employed the same hashtags both critically and supportively (\#MAGA), and revived and created new hashtags (\#BelieveSurvivors, \#WhyIDidntReport) over the one year period in response to pivotal moments. We also discuss the implications of our study for digital activism design and research including computing tools that may help under-represented voices gain visibility. Because our analysis employed automated demographic inference methods, we include a robust discussion on the ethical, privacy, and fairness implications of these tools for our analysis.  

\subsection{Ethics, Fairness, and Privacy}
Given the sensitive nature of this study, we reflect on the ethical, fairness, and privacy implications of our work. 

First, our study was exempted from review from our respective institutional review boards under 45 CFR § 46.104. Although the tweets in our study are public, to prevent re-identification outside of a tweet's intended audience, we adhere to data protections \cite{Benton:2017lq} including modifying quotes of non-organizations to avoid reverse identification \cite{Ayers_NPJDM_2018}. We only include modified tweets to contextualize our computational results. Some of the content is sensitive, so we suggest caution for readers.

Second, several recent studies have exposed a lack of fairness for demographic inference methods applied to images \cite{buolamwini2018gender} and text \cite{wooddoughty2020using,blodgett2017racial}. We acknowledge the limitations of the demographic inference methods employed in this study and discuss these in upfront in this paper (see \S\ref{sec:background}). Third, social media users have raised concerns about protecting autonomy from automatic demographic inference and suggest that increased transparency may help mediate a lack of trust \cite{Hamidi_gender_2018}. We hope that our upfront transparency about the demographic inference methods we employed combined with the data protection methods described above can help ameliorate concerns about using these tools in this specific context. 

We believe that this analysis presents an opportunity to examine large-scale patterns of representation in this online movement. While we do not want to draw final conclusions from our analysis given the limitations of the demographic inference methods that we employ, we hope that our results provide sufficient evidence to warrant further study including those inclusive of additional intersectional identities (e.g., non-binary gender, socioeconomic status, immigration status) that likely play a role in how people participate in digital activism.

Finally, the coauthors of this paper hold different racial, ethnic, and gender identities, which has helped to ensure thoughtful perspectives in the discussion of our work. However, we acknowledge that among the coauthors, we do not hold the identity of a woman of color---a critical voice in the Me Too and feminist movements.

\section{Background}\label{sec:background}
\subsection{Hashtag Activism}
Hashtags, initially introduced by Chris Messina in 2007 as a way to designate groups on Twitter, have evolved into a way to categorize conversations, find like-minded people, and even become self-referential in everyday conversations \cite{Jackson_hashtag_activism_2020}. While hashtags can amplify the reach of voices that were already mainstream in the public discourse (e.g., celebrities, those affiliated with elite institutions), hashtags now allow other voices that were formerly excluded and ignored to be heard in public discourse. Ordinary people can democratically share their narratives and engage with like-minded people rather than waiting on those with power and privilege to narrate.

Hashtag activism is a form of digital activism that uses a hashtag/hashtags to anchor the conversation(s) related to an online movement. This bottom-up approach affords a ``crowd-sourced framing process \cite{Dimon_2013_CSCW}'' that was has traditionally been held by only the leaders of a social movement. This activism has extended beyond the online realm and into offline action (i.e., computer-supported collective action) by serving as a medium of collaborative communication to mobilize people, build support for resources, organize offline actions, and warn against danger \cite{Spier_collectiveaction,McCafferty_ACM_2011}. For example, the Occupy movement assembled support for and built momentum for an offline movement \cite{Gleason_ABS_2013}. The Arab Spring uprising used hashtag activism to organize protests and avoid state actors \cite{Bruns_ABS_2013}. Hashtag activism was used to build support for public health safety measures (e.g., promoting social distancing during the COVID-19 pandemic by ``flattening the curve''). Most recently, hashtag activism was used to share individual narratives of individual- and state-sanctioned violence to coalesce into collective storytelling and renewed world-wide support for the Black Lives Matter movement \cite{Freelon_blacklives_2016}.

Previous work has examined the role of collaborative communication technologies in digital activism including how the collaborative nature of Wikipedia functions as a knowledge building tool and collective memory of a collective action \cite{Twyman_2017_black}, how singular mentions of a hashtag on social media do indeed contribute to collective storytelling (as opposed to being considered spam) \cite{Simpson_CSCW_2018}, and how offline activism groups have used collaborative communication technologies to democratically engage with the public \cite{Asad_2015_CSCW}. We extend on these previous examinations of the role of collaborative communication technologies in digital activism to understand how intersectional identities are using collaborative communication technologies to engage in storytelling, effectively framing their perspective, in a hashtag activism movement. 

\subsection{\#MeToo}
Despite that 81\% of U.S.\ women have experienced some form of sexual harassment or violence in their lifetime \cite{harassmentstats}, gendered harassment and violence have historically been framed as individual problems (e.g., victim blaming to normalize harassment and violence). In part, this has served to keep stories and subsequently the scale of the social problem hidden \cite{Kilmartin_sexualassault_2001,Meyers_blame_1996}. Feminist hashtag activism opened the door to challenging mainstream narratives about gendered violence online. In these spaces, women have shared their stories and found solidarity \cite{Williams_FemMediaStudies_2015}.

The Me Too movement was an activism movement led by women of color long before (more than 10 years before) it was endorsed by a white celebrity woman. This both increased visibility of the online \#MeToo movement, but also nearly erased the history of it; many people believe the movement was started by Milano's tweet. The frequency of storytelling in the \#MeToo social media postings by women across geographic areas, races, ethnicities, socioeconomic statuses, and political affiliations shifted the cultural narrative of gendered violence from being an individual problem to being a legitimate social problem. The collective storytelling of the movement has been credited with eliciting collective action and real change offline, ranging from removal of those committing such misconduct from positions of power in their respective institutions to increased consciousness of and interest in the prevention of misconduct \cite{Ayers_JAMAIM_2018}.

However, the \#MeToo movement has also received criticism for initially ignoring the contributions from women of color in combating gendered violence and the movement not being centered on the marginalized voices that are more vulnerable to sexual harassment and violence, exemplifying how women of color have been historically excluded in feminist movements in the U.S.\  \cite{intersectionalmetoo2,trott2020networked,leung2019metoointersectional,bowman2020maximizingdiss}. For example, the credibility of women of color is discounted in investigations and prosecutions of sexual violence \cite{Tuerkheimer_2017,Lonsway_ViolenceAgainstWomen_2012}, and women of color are subjected not only to gender bias (e.g., myth rapes \cite{Page_FemCrim_2010}), but also racial bias (e.g., the intersected racial-gendered stereotype of an ``angry black woman'') \cite{Smith_1999}.

In turn, scholars have analyzed how demographic groups contributed to the \#MeToo movement. PettyJohn et al. (2019) qualitatively examined the content of one day of \#HowIWillChange tweets (3,182 tweets), initiated to engage men, during the first week of \#MeToo \cite{pettyjohn2019howiwillchange}. The researchers found that these tweets either expressed allyship by discussing strategies to dismantle rape culture (e.g., examining their role in toxic masculinty) or resistance, including open hostility (e.g., accusing men that were allies of being weak). Hosterman et al. (2018) examined the types of social support expressed by men and women (identified by manual review of their profile) in 2,782 tweets during the first six months of \#MeToo \cite{hosterman2018twitter}. The researchers found that informational support was the most common type of support expressed by both men and women although, as a percentage of all tweets, men shared slightly more informational support compared to women who shared slightly more emotional support. Modrek and Chakalov (2019) examined the frequency of disclosures of sexual abuse or assault expressed by singular identities (gender and, separately, race/ethnicity that were inferred using a commercial marketing software) in 11,935 tweets during the first week of \#MeToo \cite{Modrek_JMIR_2019}. The researchers found that 11\% of the tweets disclosed an incidence and these tweets were overwhelmingly authored by inferred white (89\%) and female (90\%) posters.

We extend prior work by examining intersectional identities of gender and race or ethnicity---an important lens \cite{trott2020networked} for viewing this movement---in a large sample (660,237 tweets from 256,650 unique users) of tweets during the first year of \#MeToo. Rather than \textit{a priori} examining a specific issue, our work uses topic modeling to broadly examine the content, or storytelling, of the tweets.

\subsection{Demographic Inference on Social Media}
Numerous recent studies have used publicly available social media data to investigate public health matters \cite{sinnenberg2017twitter,paul2011tweet,paul2017social,gun_analysis}. However, social media posts in themselves, including tweets, present little contextualizing information. Thus, to align social media data with traditional data sources, demographic information can be useful (and is often essential for more controlled analyses). Demographics---in our case, gender, race, and ethnicity---have been used in a variety of studies to contextualize online conversations \cite{mandel2012demographic,mitchell2013geography,huang2017examining,carpenter2017real,huang2019can,Amir:2019lq}. Prior to the advent of demographic inference, data accessibility of demographics was a major hindrance to demographically-informed social media research \citep{ayers2014medicine}. 

Numerous approaches exist for demographic inference of social media including text-based \citep{rao2010demographic,demographer,cesare2017detection,volkova2013demographic,mislove2011demographic,burger2011demographic} and image-based approaches \cite{makinen2008evaluation,ng2015review}. It is widely acknowledged that both of these approaches have inherent biases and limitations. For example, known biases include, but are not limited to, a tendency for image-based approaches to incorrectly classify darker skin tones, especially for women \cite{buolamwini2018gender}, and for text-based approaches to display racial disparities \cite{blodgett2017racial}. Methods that are reliant on previous postings or social networks are potentially subject to insufficient historical data or prohibitively expensive calls to an API for a user's historical data \cite{neuraldemographer}. Additionally, all of these approaches limit the inference of gender, race, and ethnicity to a binary classification (e.g., users are either male or female and users are limited to a singular racial/ethnic category). This oversimplication overlooks the complexity of identities. First, binary classification of race and ethnicity overlooks that multiracial/ethnic identities can be comprised of a primary and secondary identity. Second, these approaches are incapable of identifying non-binary genders. Third, inferred identities may not accurately reflect their self-identified demographics. Finally, identities can be contextual shifting across situations and settings and these approaches cannot capture that nuance. 

We now upfront discuss limitations specific to the demographic inference methods employed in this study (details of the method itself can be found in the Methods section) and the potential impacts to our analysis. First, for gender, we employ a model that examines user names. The model achieves high to reasonable precision and recall for women and men, in part because of its reliance on character-level representations in text, which can identify features that are likely distinctive of gender (e.g., usage of a male vs female emoji). Nonetheless, its training data is limited to binary gender labels, and thus our model will misrepresent users who do not identify in binary terms. Additionally, our model may be biased by cultural naming norms, as its training data over-represents users in the United States. Given the high to reasonable performance, we suspect these biases do not impact the overall results of this study on a large scale. Second, for race or ethnicity, we employ an ensemble of two models: one based on unigrams of a user's most recent tweets, and the other based on known audience demographics of popular Twitter accounts (i.e., websites like ESPN) that the user follows. This ensemble method results in high performance for white users and moderate performance for people of color. To overcome this lackluster performance, we (1) favor precision (i.e., correctly identify the race or ethnicity) over recall (i.e., recalling all instances of a race or ethnicity) and (2) only label race or ethnicity if both of these models are in agreement. Similar to the gender model, examining word-level content does allow us to pick up on key features potentially indicative of race (e.g., skin tone of emoji). We suspect that although we are not recovering every post authored by people of color, we are reasonably sure that our label is correct when we do. Both of the models do not randomly misclassify gender, race, or ethnicity, but instead misclassify on data that are more ambiguous and do not contain key identifying features. We suspect that this correlates with how users' identities are \textit{perceived} on Twitter. For example, usage of skin tone and gendered emojis may also cause a human reader to infer the same identify of a user. 

We acknowledge that these limitations and biases make our study imperfect; however, we believe that the ability to understand these patterns at a large-scale level warrants their usage despite these concerns. Our study investigates these conversations across intersectional identities over a longer time period than previous work, allowing for a more nuanced view of how demographic groups engaged in this event of hashtag activism.

\section{Data and Methods}

\subsection{Data}
We retrieved one year of public tweets in English that contained the hashtags `\#metoo', `\#survivor', `\#domesticviolence', `\#abuse', `\#whyididn(o)t', and `\#myvoice' (as well as hashtags containing any of those character sequences as sub-strings) from October 1, 2017 to October 28, 2018 from the  1\% Streaming Twitter API.\footnote{\url{https://developer.twitter.com/en/docs/tweets/sample-realtime/}} These hashtags were selected based on a survey of media reports covering the movement. Our dataset consists of 660,237 tweets from 256,650 unique users.\footnote{Extrapolating from the 1\% feed, this suggests that there were 66 million English Me Too related tweets in the first year.} 

\subsection{RQ1: Who actively tweeted?}
To infer gender (female, male) for the users in our dataset, we employed NeuralGenderDemographer---a convolutional neural network that classifies the binary gender of the user---from Demographer \cite{neuraldemographer}. The model is trained on the characters of 58,046 Twitter usernames with gender identified through linked blogs. We evaluated the model on a gold-standard dataset consisting of Twitter users who explicitly self-reported their gender in a survey \cite{preoctiuc2015studying}, finding the model achieves 90.9\% precision and 83.2\% recall for women, and 71.4\% precision and 83.5\% recall for men on this dataset.\footnote{Precision refers to the proportion of predictions for a given category that were correct (i.e., exactness/quality), whereas recall refers to the number of members of a given category that were correctly classified (i.e., completeness/quantity).} The benefit of this character-level neural model is that it can learn patterns in character sub-sequences, which is beneficial for Twitter user names that contain special characters (like emojis) that may reveal a user's gender. More specific details of the model can be found in Wood-Doughty et al. (2018) \cite{neuraldemographer}. 

To infer a \textit{singular} race or ethnicity of the users in our dataset, we employ an ensemble of two methods. The first model, the EthSelfReportNeuralDemographer from Demographer \cite{wooddoughty2020using}, is a regularized logistic regression classifier that uses unigrams from the user's most recent 200 tweets to classify a user's race or ethnicity. The EthSelfReportNeuralDemographer model was trained on an automatically-collected corpus of users who self-reported their race or ethnicity in their Twitter user profile description. The second model, from Culotta et al. (2016) \cite{ethnicityfollowers2}, is a regularized logistic regression classifier that infers a user's race or ethnicity by analyzing which popular Twitter accounts that the user follows (i.e., high traffic websites with known measurements of audience demographics that visit those website). Due to limitations in available training data, both models limit classifications to the most populous racial and ethnic groups in the U.S.: Asian, black, Hispanic, and non-Hispanic white. To avoid drawing conclusions from incorrectly-labeled demographics, we favor precision (i.e., correctly identifying the race/ethnicity) over recall (i.e., identifying all instances of a given race/ethnicity) by only labeling the race/ethnicity for users for which our two classifiers agreed; if our two classifiers did not agree, we labeled race/ethnicity as `unknown'. We evaluated this ensemble method on a dataset of self-reported race/ethnicity data from Twitter users \cite{wooddoughty2020using} achieving the following precisions and recalls: 20.0\% precision and 32.4\% recall for Asian users, 50.8\% precision and 81.3\% recall for black users, 52.2\% precision and 19.4\% recall for Hispanic users, and 93.3\% precision and 87.3\% recall for white users.

\subsection{RQ2: What stories were shared?}
We employed Latent Dirichlet allocation (LDA) \cite{lda}, a probabilistic topic modeling method that infers topics from distributions of words. This method identifies salient topics in a set of documents---in this case, the tweets in our dataset. Topic models have been used to discover salient points of discussion and to delineate different foci of conversations and documents in many domains, including in social media data \cite{lau2012line,hong2010empirical,yang2014large,ramage2010characterizing,topics_socmedia,gun_analysis,discoveringhealthtopics}.

We tokenized tweets using TweetTokenizer from NLTK \cite{nltk}, removed capitalization, and removed tweets with three or fewer words. We trained LDA topic models with 20, 30, 40, and 50 topics with a vocabulary consisting of the 20,000 most frequent tokens in our dataset. We ran the models for 2,000 iterations with the hyperparameter defaults defined in Paul and Dredze (2015) \cite{sprite}\footnote{\url{https://bitbucket.org/adrianbenton/sprite/}} and averaged the final 100 iterations to estimate the LDA parameters. Upon qualitative review of the topics of each of the topic models by the authors, we selected to move forward with the topic model with 40 topics because it produced the most qualitatively meaningful and cohesive topics. Then we assigned final labels for the topics by qualitatively reviewing the word distributions of the topic with randomly selected tweet examples. We excluded topics from our results that did not have any coherent semantic groupings (hence why we present fewer than 40 topics in the results); non-coherent groupings were found by locating topics whose top-10 tokens did not frequently co-occur in the same documents (as measured by per-topic normalized pointwise mutual information), as well as topics whose tokens were primarily function words (e.g., determiners, prepositions) and not content words. 

We describe our results in three ways. First, by the topic labels, words associated with each topic, and probability of each topic's occurrence in the dataset. The probability of a topic was calculated by counting the number of tokens occurring in the tweet dataset which are members of that topic, then normalizing by the number of tokens which occur in \textit{any} topic in the dataset. This method allows us to account for tweets which contain discussions of multiple topics, and ensures that the probability of all topics will sum to 1. Second, by temporally examining the most probable topics during spikes in tweet volume. Third, we compare topics across intersectional demographic groups using both conditional probabilities and pointwise mutual information (PMI). The conditional probability of a topic characterizes the frequency of the topic by demographic group; this is calculated in the same way as overall topic probabilities, but here we restrict the dataset to tweets from a particular demographic group. In contrast, PMI characterizes how distinctive a topic is for each demographic group. Formally, PMI measures the deviation of the joint distribution ($x$,$y$) from the separate independent distributions of $x$ and $y$. Here, $x=$~p(demographic) and $y=$~p(topic), resulting in a metric which roughly captures how distinctive a given topic is of a given demographic group.

\begin{align*}
    \text{PMI}(x;y) = \log\frac{p(x,y)}{p(x)p(y)}
\end{align*}

Finally, we provide modified quotes that exemplify the topics to contextualize the qualitative narrative of the topic. 

\subsection{RQ3: What hashtags were associated?}
Similarly to our topic analysis, we examine the overall probability of specific hashtags in our dataset. We then analyze and describe which hashtags are most commonly used among each intersectional demographic group using conditional probabilities, and which hashtags are more distinctive of each demographic group using PMI. We also examine the hashtags temporally in a similar nature to how we temporally investigate topics.

\begin{table}[t]
    \centering
    \resizebox{0.57\columnwidth}{!}{
    \begin{tabular}{lcccc}
        \toprule
        Demographic & Users & Percentage & Mean & Stdev \\
        \midrule
        Women & 133,717 & 53.32\% &  2.85 & 19.01 \\
        Men & 119,432 & 46.68\% & 2.38 & 19.64 \\
        \hline
        White & 99,601 & 59.65\% & 2.90 & 22.77 \\
        Black & 42,164 & 25.25\% & 2.53 & 14.98 \\
        Asian & 16,422 & 9.84\% & 2.36 & 25.36 \\
        Hispanic & 8,781 & 5.26\% & 2.00 & 8.31 \\
        \midrule
        White Women & 52,560 & 31.48\% & 3.24 & 21.67 \\
        White Men & 47,041 & 28.17\% & 2.52 & 23.93 \\
        Black Women & 22,397 & 13.41\% & 2.74 & 15.64\\
        Black Men & 19,767 & 11.84\% & 2.28 & 14.19\\
        Asian Women & 8,561 & 5.12\% & 2.41 & 24.49 \\
        Asian Men & 7,861 & 4.71\% & 2.31 & 26.27 \\
        Hispanic Women & 5,055 & 3.03\% & 2.06 & 7.48 \\
        Hispanic Men & 3,726 & 2.23\% & 1.93 & 9.32 \\
        \bottomrule
    \end{tabular}}
    \caption{The number of users, percentage of classifiable users, mean authored tweets, and standard deviation of authored tweets for each demographic identity.} %For the intersectional statistics, we collapse non-white and Hispanic users into the people of color (POC) category for higher precision in our statistics.}
    \label{n_tweets}
\end{table}

\section{Results}
\subsection{RQ1: Who actively tweeted?}
Table~\ref{n_tweets} presents the number of users per-demographic, as well as the mean and standard deviation of tweets per-user across the inferred demographic groups (gender, race, and ethnicity).\footnote{11,176 users were classified as unknown by our gender classification model. These tweets were included during topic modeling and overall hashtag probability calculations, but are otherwise excluded from analysis.} Approximately 52\% of Twitter users are women,\footnote{As of Jan. 2018, 23\% of men and 24\% of women in the U.S. use Twitter \cite{socialmediastats}. The most recent U.S. census states that 50.9\% of the U.S. population are women, whereas 49.1\% are men \cite{us_census}. We make a simplifying assumption that Twitter users are either men or women, and use these proportions to calculate that approximately 52\% of Twitter users are women and 48\% are men.} whereas women represented 53.32\% of users in our dataset. Women also authored significantly more \#MeToo tweets per-user (M=2.85, SD=19.01) than men (M=2.38, SD=19.64) ($p < .0001$).

In our dataset, the largest racial group is non-Hispanic white, comprising approximately 60\% of users inferred; compare to the U.S.\ population which is comprised of approximately 61\% non-Hispanic white individuals.\footnote{\url{https://www.census.gov/quickfacts/fact/table/US/PST045218}} Precision and recall for our ensemble method for the non-Hispanic white group was high (see Methods). Black users comprised 25\% of our dataset, followed by Asian (10\% of users), and Hispanic (5\% of users). Our ensemble method achieves significantly lower precision and recall for people of color; thus, there are likely more users that participated in the \#MeToo movement that were not recalled. We opted to favor precision over recall and use an ensemble method where both classifiers had to agree on a label which, in practice, likely translates to labels that would be consistent with how other users perceive racial or ethnic identity. There were significant differences in the number of tweets per-user between racial and ethnic groups ($p<.001$ in a one-way ANOVA test). Specifically, post-hoc analyses using Tukey's HSD \cite{tukeytest} indicated that white users authored significantly more \#MeToo tweets per-user (M=2.90, SD=22.77) than people of color---Asian (M=2.36, SD=25.36), black (M=2.53, SD=14.98), or Hispanic (M=2.00, SD=8.31) users ($p<.01$). Differences in tweets per-user between Asian, black, and Hispanic groups were not significant ($p>.1$).

Given the differences between the tweeting patterns across gender, race, and ethnicity, it is unsurprising that the number of tweets per-user significantly differed between intersectional demographics as well ($p<.001$ in an ANOVA test). Specifically, the number of tweets per-user was significantly higher for white women (M=3.24, SD=21.67) than any other intersectional demographic ($p<.001$ in a series of Tukey \emph{post-hoc} tests), while differences between any other two intersectional demographics was not significant ($p>.1$ in a series of Tukey \emph{post-hoc} tests). This suggests that, despite the Me Too movement's origins among women of color, the viral Twitter \#MeToo movement was primarily represented by white women.

\subsection{RQ2: What stories were shared?}
\subsubsection{Topics in the General Dataset}
Table~\ref{fulltopics} presents the topics identified by the topic model, as well as the probability of occurrence in our dataset and representative tokens for each topic.\footnote{The `wrestling/sports' topic is from the `survivor' hashtags: \#SurvivorSeries refers to a wrestling event, and \#SurvivorFinale refers to the game show Survivor. The `Animals' rights' topic results from our search for `abuse' hashtags, for we obtain \#AnimalAbuse and \#Abuse (the latter of which is often applied to animal abuse discussions). We removed any tweet in our dataset with the \#SurvivorSeries, \#SurvivorAU, \#SurvivorFinale, and \#AnimalAbuse hashtags before running all other quantitative analysis scripts or generating figures.}

\begin{table*}[t]
    \centering
    \resizebox{\linewidth}{!}{
    \begin{tabular}{p{3.35cm}cp{5cm}||p{3.35cm}cp{5cm}}
    \toprule
         \centering Topic & Prob. & Representative Tokens & \centering Topic & Prob. & Representative Tokens  \\
    \midrule
         \centering Love/support & .044 & love good happy thank like family great best & \centering Hollywood accusations/Weinstein & .032 & weinstein hollywood harvey star accused alyssamilano porn woman \\\hline
         \centering Trump insults & .043 & realdonaldtrump f***ing f*** s*** trumprussia hell lying stupid man & \centering Domestic violence/abuse survivors & .031 & abuse violence domesticviolence survivors survivor awareness domestic victims trauma \\\hline
         \centering MAGA & .041 & metoo resist maga theresistance fbr resistance realdonaldtrump fbrparty qanon trumpshutdown & \centering Rape culture & .030 & sexual assault harassment abuse rape victims violence culture predator misconduct \\\hline
         \centering Article/thread sharing & .040 & read thread story media article children body metoo piece & \centering Wrestling/sports (unrelated) & .029 & survivorseries wwe team raw vs nfl sdlive match players play \\\hline
         \centering Sexual allegations & .039 & sexual trump metoo assault women accused everyone telling congress allegations resign guilty & \centering Animals' rights (unrelated) & .029 & protectwildlife sign animals trophy animal petition dog hunting stop kill \\\hline
         \centering Women's rights & .039 & women rights human right justice due political civil means & \centering Social media & .028 & twitter tweet account facebook tweets accounts video followers social posted block \\\hline
         \centering Promoting women's stories & .037 & metoo women thank stories survivors speak story forward sharing believe brave support truth & \centering Racism & .028 & white black people sarah sanders hate serious racism racist face \\\hline
         \centering U.S. president & .036 & trump president america obama country american people history party united states & \centering National security & .027 & trump russia putin president security donald north korea saudi national russian \\\hline
         \centering Gender \& sexual violence & .035 & sexually women assaulted men man mean reminder boys people sexual abused & \centering Elections/voting & .026 & vote election gop state democrats republicans house democratic candidate running november voters voting \\\hline
         \centering \#MeToo movement & .035 & metoo movement women time timesup silence moment campaign taranaburke started hollywood & \centering Politics in the media & .026 & news fox cnn john msnbc president press white rally media kazweida fake kelly \\\hline
         \centering Democrat politics & .034 & clinton hillary bill obama 2018 bluesea1964 party democrats hillaryclinton & \centering Kavanaugh hearings & .024 & kavanaugh ford dr court brett supreme judge senate blasey christine believe believesurvivors senatorcollins fbi hearing jeffflake \\\hline
         \centering Republican politics & .034 & realdonaldtrump potus gop foxnews speakerryan tedlieu senatemajldr donaldtrumpjr seanhannity senategop & \centering Police brutality & .021 & police state arrested officer killed black woman news video copcrisis \\\hline
         \centering News/media sharing & .033 & report video story youtube hero abuser change via & \centering Mueller investigation & .021 & trump fbi mueller investigation cohen michael breaking russia campaign manafort comey attorney sessions lawyer doj \\\hline
         \centering Legislative discussion & .033 & tax million money pay bill gop cut americans health care ryan billion & \centering Child abuse & .020 & children sex child woman abuse rape parents wife kids matter families friend body church girlfriend family forcing \\\hline
         \centering Sharing personal assault stories & .032 & years ago old told raped days whyididntreport months man took today said last & \centering School shootings & .015 & school gun nra high students shooting moore guns kids parkland violence mass student schools florida teachers davidhogg111 \\
         \bottomrule
    \end{tabular}
    }
    \caption{The topics discovered by topic modeling, probabilities of each topic in the dataset, and most probable tokens associated with each topic.}
    \label{fulltopics}
\end{table*}

The topics covered a range of issues including disclosures of sexual harassment and violence, political discourse, supportive discourse, and discussions of public figures involved in high-profile sexual harassment and violence investigations. Below, we describe the ten most probable topics in the general dataset.

The most probable topic in the general dataset was `love/support'. These tweets often expressed solidarity and emotional support for those who had experienced sexual harassment or violence and were participating in the online movement. Often these tweets were directed towards people who had disclosed an experience of sexual harassment or violence. For example, 
\begin{quote}
    \textit{``love you girl! thanks for sharing \#theresistance''}
\end{quote}

Of the ten most probable topics, three topics (`Trump insults' (second most probable), `MAGA' (third most probable), `U.S.\ president' (eighth most probable)) focused explicitly on President Trump, each with different emphasis. The `Trump insults' topic consisted of tweets that primarily expressed criticisms, complaints, and jokes aimed at Trump. For example, 

\begin{quote}
    \textit{``I hate everything about this administration. HATE. EVERYTHING. F*** Trump, and f*** you if you voted for this''}
\end{quote}

The `MAGA' topic featured mentions of President Trump's campaign slogan of Make America Great Again. This topic initially appeared as if it would be a topic opposed to the \#MeToo movement, but we found that this abbreviation was often employed subversively and ironically to call attention to a high-profile figure who has been accused, but not investigated, of sexual harassment and assault. For example, 

\begin{quote}
    \textit{``i believe the women that have accused @realDonaldTrump of sexual assault \#DumpTrump \#MAGA \#TheResistance''}
\end{quote}

\begin{quote}
    \textit{``If you really want to \#MAGA, start speaking up against these guys who abuse their power and \#BelieveSurvivors.''}
\end{quote}

The `U.S.\ president' topic included tweets that expressed discussions of the presidency in general, including discussions of former U.S.\ presidents (primarily Barack Obama and George W.\ Bush). For example, 

\begin{quote}
    \textit{``@realDonaldTrump -- you are causing the divisiveness. At least when the Bush men were POTUS I respected them because they were doing SOMETHING to move America forward.''}
\end{quote}

The fourth most probable topic, `Article/thread sharing', included tweets that primarily focused on providing informational support by sharing data, like statistics on the prevalence of sexual harassment and assault, and news including reactions to the news. For example,

\begin{quote}
    \textit{``Janelle Monae delivered an empowering message at the Grammys about the \#MeToo movement and \#TimesUp campaign. Read the full speech: <link>''}
\end{quote}

\begin{quote}
    \textit{``Important piece on the deep-rooted discrimination that women of color experience from the justice system as \#MeToo survivors <link>''}
\end{quote}

The fifth most probable topic, `sexual allegations', included tweets that primarily focused on allegations and investigations of sexual harassment and violence against high-profile individuals. For example,

\begin{quote}
    \textit{``The 16 women who accused Trump of sexual assault are telling their story in one video---please share!''}
\end{quote}

The sixth most probable topic of `Women's rights' included tweets that focused on historical and current feminist movements including women's suffrage, reproductive rights, and---most relevantly---sexual assault. For example,

\begin{quote}
    \textit{``ofc the war on women keeps goin. how much longer we gonna have to fight like we did to vote? \#believewomen''}
\end{quote}

Similar to the most probable topic in the dataset ('love/support'), the seventh most probable topic in the general dataset, `promoting women's stories', included tweets that explicitly provided emotional support for people who disclosed incidences of sexual harassment or violence either publicly or privately. For example, 
\begin{quote}
    \textit{``To all the women sharing stories of sexual assault and sexual harrassment, you are not alone. Thank you for your bravery. \#metoo''}
\end{quote}

Finally, the ninth and tenth most probable topics of `gender and sexual violence' and `\#MeToo movement' focused on discussions of gendered violence and how it is reinforced through cultural norms. The former often entailed sharing personal experiences or opinions about the extent and nature of sexual violence against women, as well as broader discussions of the causes that may have led to the \#MeToo movement. The latter tended to include direct discussions of developments in the \#MeToo movement, including personal \#MeToo stories as well as articles specifically about the movement. For example, 

\begin{quote}
    \textit{``fr, stop teaching women to not get assaulted. Teach boys and men not to assault''}
\end{quote}

\begin{quote}
    \textit{``Evangelical women just joined \#Metoo, and They're Urging Churches to Address Abuse''}
\end{quote}

Although not the most probable topics, notable topics including  `racism' and police brutality' were present. Tweets that contained these topics featured discussions of institutional and systemic racism and bias that contribute to the increased vulnerability of women of color to sexual harassment and violence. These topics highlighted differential racial justice for victims of sexual harassment and violence in our criminal system while contextualizing the \#MeToo movement with respect to the disparities in treatment across racial and ethnic identities. For example, 

\begin{quote}
    \textit{``It's cause we got to where we are because men, specifically white men, have been given most of the power in this country.''}
\end{quote}

\begin{quote}
    \textit{````Whining'' on Twitter helped us with targeted activism on two important fronts: police brutality and sexual assault. Neither would be a conversation without hashtags like \#MeToo and \#BLM. Bodycams and firing powerful men would not have been possible.''}
\end{quote}

\begin{figure}[t]
    \centering
    \includegraphics[scale=0.65]{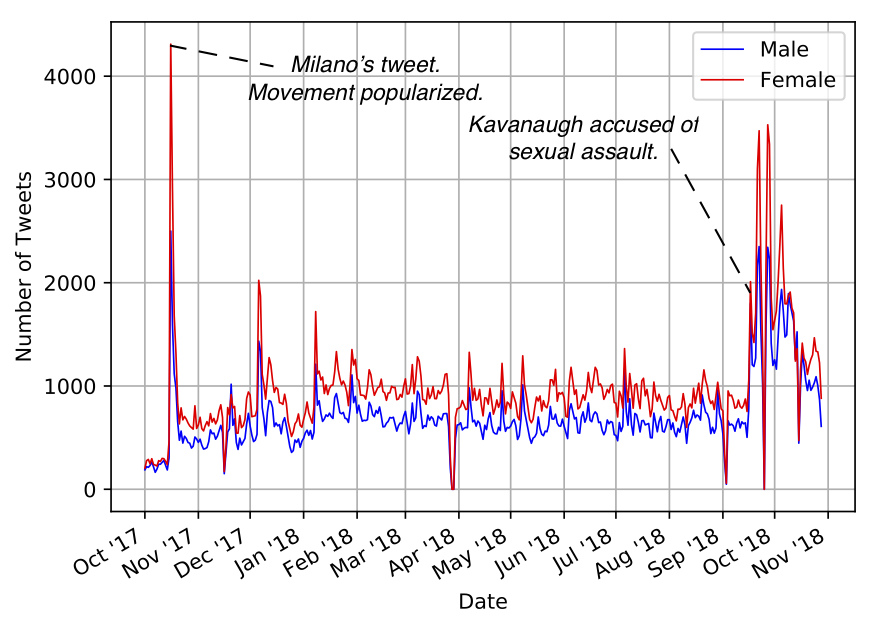}
    \caption{Daily tweet volume by gender in our dataset.}
    \label{tweetvolume}
\end{figure}

\subsubsection{Variations in Topics Temporally}
Figure~\ref{tweetvolume} presents the daily number of tweets by the gender of the tweet author. The start of the viral movement is clear on October 15, 2017. Approximately 68\% of all original tweets---not retweets or replies---in the first 48 hours of the movement were authored by women (whereas 53\% of all original tweets in the first year of the movement were authored by women.) The most probable topics during initial popularity are `promoting women's stories' ($P=.077$), `rape culture' ($P=.075$), `gender and sexual violence' ($P=.075$), and `Sexual allegations' ($P=.067$). These topics primarily focus on disclosures of personal experiences of sexual harassment or violence, providing emotional support for those who have experienced sexual harassment or violence, and sharing informational support including data about prevalence of sexual harassment and violence. 

After the initial viral popularity of \#MeToo on Twitter, tweet volume declined and remained fairly stable at just under 2,000 tweets per day; however, there were smaller spikes in December 2017 after TIME magazine named the \#MeToo movement person of the year, and in January 2018 when Catherine Deneuve denounced the movement and later when Aziz Ansari was accused of sexual assault. There was a much larger spike in mid-to-late September 2018, when the \#MeToo hashtag was revived again during the widely publicized accusations of sexual violence against Judge Brett Kavanaugh.\footnote{Brett Kavanaugh is a U.S. Supreme Court Justice whose U.S. Senate confirmation hearings took place in September to October 2018. During the hearings, he was accused of a sexual assault that took place while in high school. The issue became of national interest in the wake of the first year of to the \#MeToo movement.} The most probable topics that were discussed during this latter spike (September 15 to October 7, 2018) included `Kavanaugh hearings', `Promoting women's stories', ($P=.057$), and `sharing personal assault stories' ($P=.047$). For example,
\begin{quote}
    \textit{``\#WhyIDidntReport The culture that discouraged Christine Blasey Ford from speaking out punishes women their entire careers.''}
\end{quote}

\begin{figure}
    \centering
    \includegraphics[width=\linewidth]{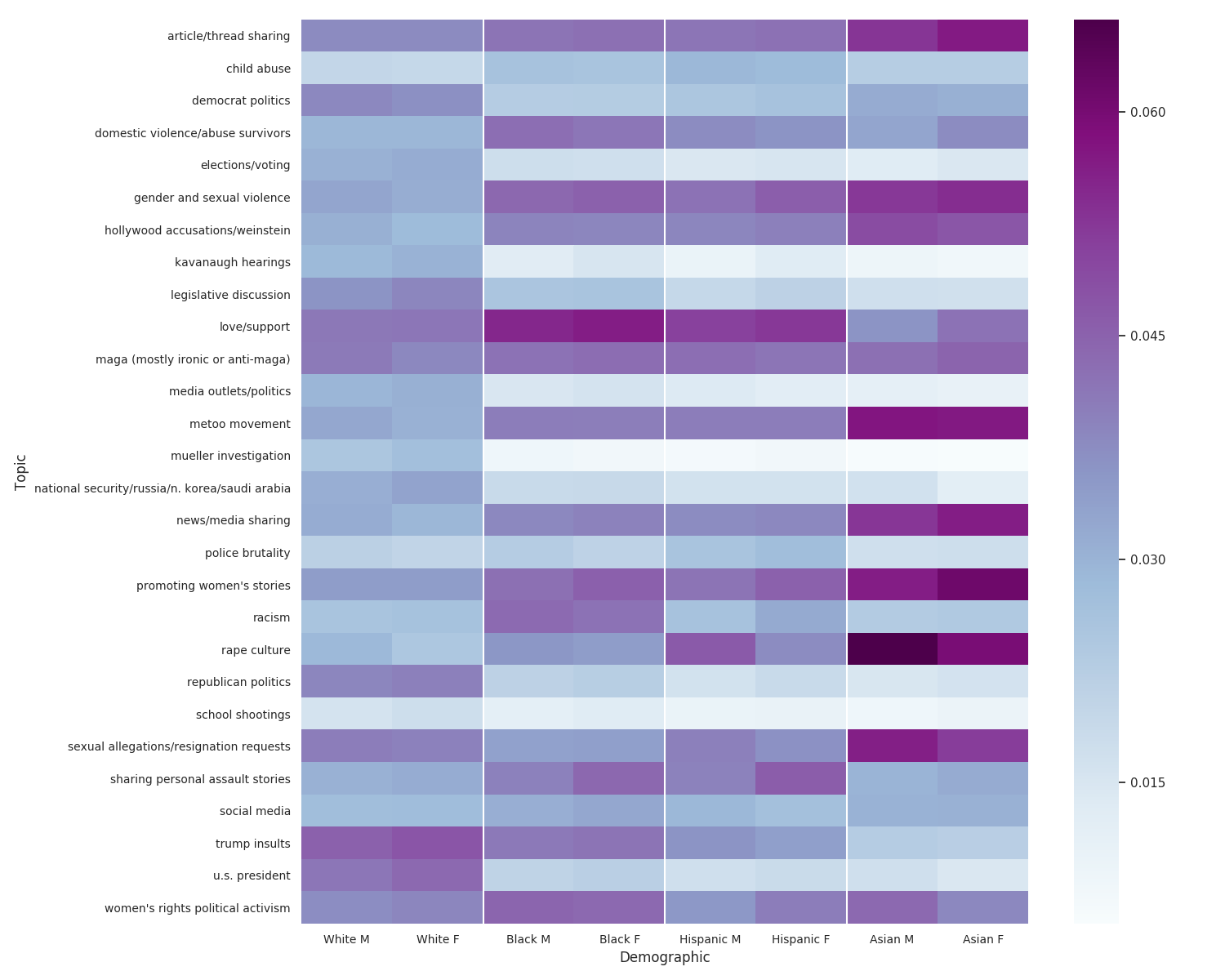}
    \caption{Conditional probabilities of topic occurrence given intersectional demographics. Demographics are grouped first by race/ethnicity then gender along the x axis, as topics seemed to vary more markedly by race/ethnicity. F = female, M = male.}
    \label{fig:topics_heatmap}
\end{figure}

\begin{table}
\resizebox{0.431\linewidth}{!}{
    \begin{tabular}{lrl}
        \multicolumn{3}{c}{Women} \\
        \toprule
        Black & PMI & \\
        \midrule
        racism & \rbar{0.546}\\
        sharing personal assault stories & \rbar{0.383} \\
        domestic violence/abuse survivors & \rbar{0.357} \\
        gender and sexual violence & \rbar{0.313} \\
        love/support & \rbar{0.300} \\
        \midrule
        White & & \\
        \midrule
        mueller investigation & \rbar{0.404} \\
        kavanaugh hearings & \rbar{0.367} \\
        u.s. president & \rbar{0.307} \\
        media outlets/politics & \rbar{0.289} \\
        national security & \rbar{0.286} \\
        \midrule
        Hispanic & & \\
        \midrule
        sharing personal assault stories & \rbar{0.446} \\
        child abuse & \rbar{0.425} \\
        police brutality & \rbar{0.340} \\
        gender and sexual violence & \rbar{0.321} \\
        rape culture & \rbar{0.253} \\
        \midrule
        Asian & & \\
        \midrule
        rape culture & \rbar{0.945} \\
        news/media sharing & \rbar{0.731} \\
        metoo movement & \rbar{0.689} \\
        promoting women's stories & \rbar{0.689} \\
        gender and sexual violence & \rbar{0.590} \\
        \bottomrule
    \end{tabular}}
    \quad
    \resizebox{0.44\linewidth}{!}{
    \begin{tabular}{lrl}
        \multicolumn{3}{c}{Men} \\
        \toprule
        Black & PMI & \\
        \midrule
        racism & \bbar{0.585}\\
        domestic violence/abuse survivors & \bbar{0.396} \\
        child abuse & \bbar{0.310} \\
        gender and sexual violence & \bbar{0.267} \\
        love/support & \bbar{0.253} \\
        \midrule
        White & & \\
        \midrule
        mueller investigation & \bbar{0.281} \\
        kavanaugh hearings & \bbar{0.277} \\
        elections/voting & \bbar{0.230} \\
        media outlets/politics & \bbar{0.228} \\
        u.s. president & \bbar{0.225} \\
        \midrule
        Hispanic & & \\
        \midrule
        rape culture & \bbar{0.553} \\
        child abuse & \bbar{0.464} \\
        police brutality & \bbar{0.247} \\
        sharing personal assault stories & \bbar{0.219} \\
        hollywood accusations/weinstein & \bbar{0.213} \\
        \midrule
        Asian & & \\
        \midrule
        rape culture & \bbar{1.089}\\
        metoo movement & \bbar{0.692} \\
        news/media sharing & \bbar{0.627} \\
        hollywood accusations/weinstein & \bbar{0.577} \\
        promoting women's stories & \bbar{0.562} \\
        \bottomrule
    \end{tabular}}
    \caption{Top five most distinctive topics for each demographic group as measured by pointwise mutual information.}
    \label{tab:topics_pmi}
\end{table}

\subsubsection{Variations in Topics Across Demographic Groups}
Figure~\ref{fig:topics_heatmap} presents the conditional probabilities of each of the topics by each demographic group (i.e., the probability of topic occurrence within a demographic group) and Table~\ref{tab:topics_pmi} presents the PMI of each of the topics by each demographic group (i.e., how distinct each topic is for each demographic group, with high PMI indicating the topic is more distinctive of that group). The tweets of all demographic groups discussed a mixture of personal (e.g., `sharing personal assault stories') and political topics (e.g., `democrat politics'). We first discuss our observations across race and ethnicity, then across gender, and finally across intersectional identities of gender, race, and ethnicity.

\textbf{Comparing Topics Across Race and Ethnicity.} There were some notable differences between topics expressed in tweets authored by people of color versus white users. For example, tweets authored by white users more often discussed a wider variety of topics (e.g., topics ranging from `Mueller investigation' to `sexual allegations/resignation requests' to `elections/voting') compared to tweets authored by people of color, which generally focused on topics that centered around social support (e.g, on `promoting women's stories', `gender and sexual violence', and `love/support'). Tweets authored by white users were slightly more likely to discuss political topics (e.g., the topics of `Republican politics' and `U.S. president') than those authored by people of color, who were slightly more likely to disclose personal incidences of sexual harassment and assault (e.g., `sharing personal assault stories'), but we reiterate that this difference is relatively small. The topics of `rape culture' and `gender and sexual violence' were more likely among tweets authored by Hispanic and Asian users, specifically those authored by males. For example,

\begin{quote}
    \textit{``Self respect is sorely needed when you have been sexually violated in our culture. Aggressors notice this weakness''} -- Hispanic Male Author
\end{quote}

Consistent with the observations above, the topic of `love/support', which focuses on providing emotional support, is more likely among tweets authored by black users. For example,
\begin{quote}
    \textit{``\#Love is important. We'll all get through \#MeToo too''} -- Black Male Author
\end{quote}

\textbf{Comparing Topics Across Gender.} We now observe topics expressed in tweets across gender. In general, the variety of topics discussed varied more markedly across race and ethnicity than it did across gender. At the surface level, this may be surprising because  the \#MeToo movement was primarily centered around women's experiences of sexual harassment and violence; thus, we expected a more noticeable divergence between topics discussed in tweets authored by females compared to males. Instead, we observe that the topics discussed in tweets are similar across gender within each race/ethnicity, but that males and females employ different language to discuss these topics (e.g., they often discuss different perspectives within a topic). Consider, for example, the following tweets from the `Kavanaugh hearings' topic:

\begin{quote}
    \textit{````I survived. I was fine.'' This is the mantra of women everywhere \#BelieveSurvivors \#StopKavanaugh''} -- White Female Author
\end{quote}

\begin{quote}
    \textit{``The drinking is important. if Brett was a churchgoer he wouldn't have a rep for binge drinking. character counts!''} -- White Male Author
\end{quote}

The former tweet (White F) contributes to the topic by sharing the pervasiveness of sexual harassment and violence, whereas the latter (White M) contributes to the topic by discussing the Kavanaugh's character. This exemplifies a noticeable difference that we observed across demographics: male authors tended to contribute to the conversation by framing the issues as a general public or political problem, whereas female authors tended to discuss them by drawing on and sharing their personal experiences.

\textbf{Comparing Topics Across Intersectional Identities.} We now observe topics expressed in tweets across intersectional identities of gender, race, and ethnicity. As shown in Table~\ref{tab:topics_pmi}, tweets authored by white users (male and female) were more likely than tweets authored by other identities to relate the \#MeToo movement to contemporary social issues and political events, such as school shootings, discussions on the voting process, and the Mueller investigation (as well as more directly related topics, such as the Kavanaugh hearings). In contrast, tweets authored by black females emphasized the broader presence of racial discrimination and differential treatment. For example,

\begin{quote}
    \textit{``Bill Cosby was always gonna go down as a \#MeToo sacrifice. They all guilty but the black men are going to jail''} -- Black Female Author
\end{quote}

\begin{quote}
    \textit{``Are you saying she should put aside her blackness for the sake of the \#MeToo movement? Are the 2 exclusive? Really?''} -- Black Female Author
\end{quote}

\begin{quote}
    \textit{``I will never understand how we can live in a time of \#MeToo, school shootings, our president being a \#felon, and there's nothing being done about any of it.''} -- White Female Author
\end{quote}

Additionally, tweets authored by black and Hispanic females tended to share personal experiences of sexual harassment and violence more often than their counterparts (white and Asian females). This is evidenced by the relatively high PMI in Table~\ref{tab:topics_pmi} for this topic among that demographic identities.

In general, tweets authored by males were more likely to discuss the accusations of sexual assault against high-profile individuals, such as Harvey Weinstein (`Hollywood accusations/Weinstein' topic). That said, the tweets authored by black males tended to discuss unfair treatment across racial and ethnic groups when they discussed Me Too, similarly to tweets authored by black female users. For example, 

\begin{quote}
    \textit{``black men always seen as rapists more than anyone else. took a whole damn hashtag to get people to notice that famous white men ain't all innocent''} -- Black Male Author
\end{quote}

In a similar vain, tweets by Hispanic men and women focused slightly more on tying contemporary issues like police brutality against bodies of color with the \#MeToo movement. For example, 

\begin{quote}
    \textit{``When is \#MeToo coming for cops? officers keep sexually assaulting undocumented women and punishing people who point out anything wrong''} -- Hispanic Male Author
\end{quote}

Notably we observed less difference in topics expressed in tweets of Asian females and males than other genders. Tweets authored by both Asian females and males expressed unification in topics including tending to share more news articles about Me Too than other demographics. For example,

\begin{quote}
    \textit{``Interesting article about all these celebrities ``fall from grace'' lately: <link>''} -- Asian Male Author
\end{quote}

\subsection{RQ3: What hashtags were associated?}
Hashtags can function as a link between a tweet and a larger conversation. Thus, in addition to analyzing the topics of the tweets in our datasets, we also investigate which hashtags are used and how they are employed across different demographic identities.

\subsubsection{Hashtags in the General Dataset}
\begin{figure}
    \centering
    \includegraphics[width=0.9\linewidth]{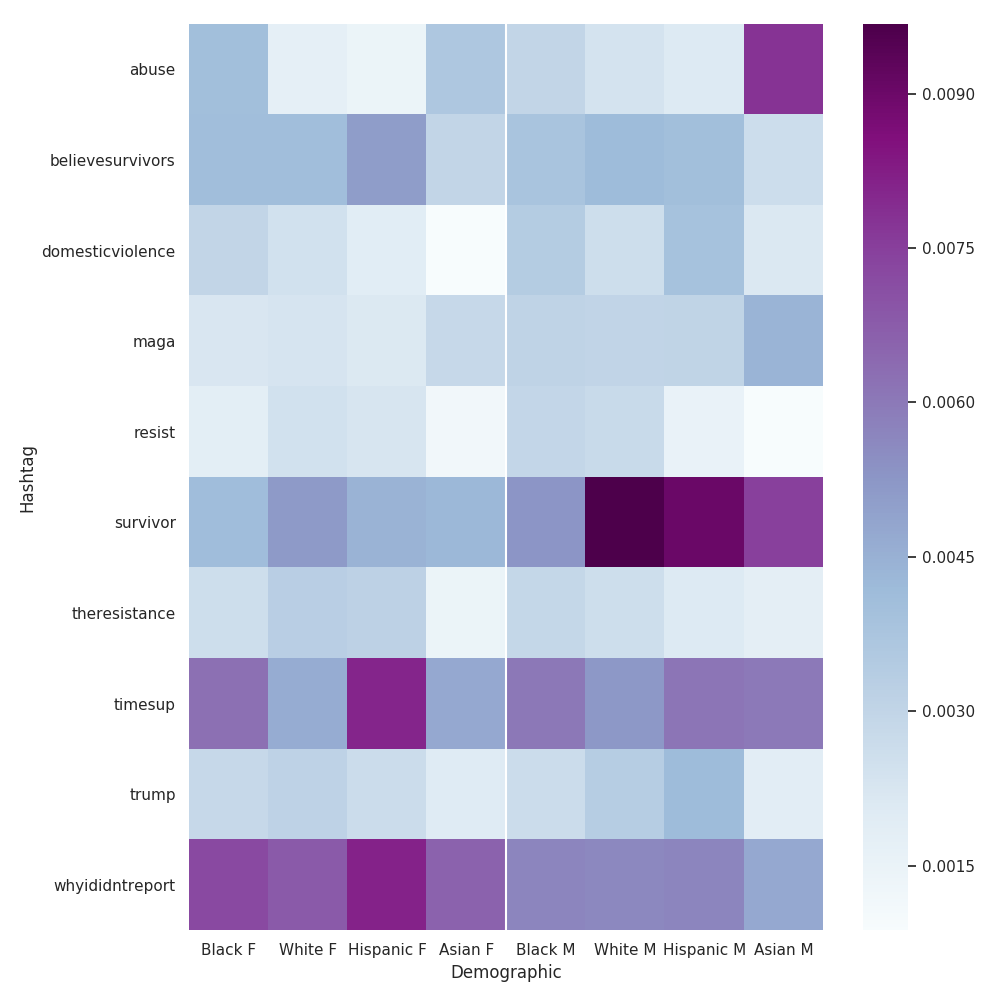}
    \caption{Conditional probabilities of the top-$10$ most probable hashtags (excluding \#MeToo) by intersectional demographic. Demographics are grouped by gender first and then race and ethnicity becauase hashtags seemed to vary more markedly by gender. F = female, M= male.}
    \label{fig:hashtags_heatmap}
\end{figure}

The most probable hashtag in our dataset was \#MeToo and the top 10 next most probable hashtags in our dataset, shown in Figure~\ref{fig:hashtags_heatmap}, include (in descending order) \#WhyIDidntReport, \#Survivor, \#BelieveSurvivors, \#TimesUp, \#abuse, \#TheResistance, \#DomesticViolence, \#Trump \#MAGA, and \#resist. 

The five most probable hashtags in our dataset---\#MeToo, \#WhyIDidntReport, \#Survivor, \#BelieveSurvivors, and \#TimesUp---each work hand-in-hand with the objectives of the \#MeToo movement, anchoring multiple perspectives (\#Survivor linking stories of surviving sexual harassment or violence and \#WhyIDidntReport linking stories of people who did not report these incidents) and simultaneous movements (\#BelieveSurvivors emphasizing historical tendencies for the burden of proof to be upon the victims of sexual harassment and violence and \#TimesUp advocating for the end of workplace sexual harassment and violence). Other probable hashtags which are applicable more broadly include \#Abuse and \#DomesticViolence.

Discussions of politics are pervasive in our dataset, and this is reflected in the next most probable hashtags. \#Trump and \#MAGA are the 8th and 9th most probable hashtags in our dataset, respectively.
We found that tweets incorporating this hashtag were frequently anti-Trump and tended to employ MAGA in a subversive way, consistent with our findings about the topic. Tweets containing \#Trump and \#MAGA often contained hashtags like \#TheResistance and \#FBRParty (Follow-Back Resistance Party), which are associated with opposition to the presidency. In a random subsample of 100 \#MAGA tweets in our dataset, 37 were ironic and/or anti-Trump, and 12 were neutral/unclear.

\subsubsection{Variations in Hashtags Temporally}
During the initial viral popularity of the \#MeToo movement in the first week after October 15, 2017, the most probable hashtags were \#MeToo, \#NoMore, \#Survivor, \#IBelieveYou, and \#HimThough. %Across demographics, these top hashtags were quite consistent. However, for people of color, we also noted the presence of \#HowIWillChange in the top 5 most probable hashtags during this initial spike. Women---especially women of color---were more likely to use the \#HimThough hashtag than other demographics. Men of color were more likely to use \#Weinstein than other demographics, and far less likely to use \#Trump.

During the latter spike related to the Kavanaugh hearings (defined here as September 15 to October 7, 2018), the top 5 most probable hashtags were \#MeToo, \#WhyIDidntReport, \#BelieveSurvivors, \#Kavanaugh, and \#StopKavanaugh. These were thematically similar to the first spike, but with the notable addition of new hashtags (\#Kavanaugh and \#StopKavanaugh) directly related to the Kavanaugh hearings and hashtags originating (\#WhyIDidntReport and \#BelieveSurvivors) in response to support Professor Christine Blasey Ford and used broadly to share incidences of sexual harassment and violence that went unreported. As with the first spike, hashtags were generally consistent across demographics---this time, even more so than during the start of the movement. Indeed, none of the demographic groups had different top-5 hashtags during this time period. Thus, this moment was perhaps the most unified with respect to the content of its conversations across demographics, as there was a singular popular event around which users could direct their discussions.

\subsubsection{Variations in Hashtags Across Demographic Groups}
Figure~\ref{fig:hashtags_heatmap} presents the conditional probabilities of the 10 most probable hashtags in our dataset expressed in tweets by each demographic group, and Table~\ref{tab:hashtags_pmi} presents the top five distinctive, as measured by PMI, hashtags expressed in tweets for each demographic group. 

\begin{table}
\resizebox{0.4\textwidth}{!}{
    \begin{tabular}{lrl}
        \multicolumn{3}{c}{Women} \\
        \toprule
        Black & PMI & \\
        \midrule
        EmbracingMyself & \rbar{2.772}\\
        Friendship & \rbar{2.423} \\
        Brutality & \rbar{2.380} \\
        Law & \rbar{2.136} \\
        Humanity & \rbar{2.089} \\
        \midrule
        White & & \\
        \midrule
        Leadership & \rbar{1.242} \\
        SexAbuseChat & \rbar{1.128} \\
        ClimateChange & \rbar{0.845} \\
        Facebook & \rbar{0.702} \\
        GenderEquality & \rbar{0.651} \\
        \midrule
        Hispanic & & \\
        \midrule
        WCW & \rbar{1.813} \\
        BreakingNews & \rbar{1.712} \\
        BoycottNRA & \rbar{1.608} \\
        NoMore & \rbar{1.494} \\
        Kavanope & \rbar{1.470} \\
        \midrule
        Asian & & \\
        \midrule
        TakePride & \rbar{3.388} \\
        VoteThemOut2018 & \rbar{2.269} \\
        Veterans & \rbar{2.066} \\
        Healing & \rbar{1.934} \\
        MosqueMetoo & \rbar{1.649} \\
        \bottomrule
    \end{tabular}}
    \quad
    \resizebox{0.51\textwidth}{!}{
    \begin{tabular}{lrl}
        \multicolumn{3}{c}{Men} \\
        \toprule
        Black & PMI & \\
        \midrule
        HumanTrafficking & \bbar{2.490}\\
        Bullying & \bbar{2.021} \\
        Suicide & \bbar{1.690} \\
        StopChildAbuse & \bbar{1.317} \\
        SexualAbuse & \bbar{1.206} \\
        \midrule
        White & & \\
        \midrule
        Pray & \bbar{1.968} \\
        Jesus & \bbar{1.758} \\
        God & \bbar{1.308} \\
        SubstanceAbuse & \bbar{1.272} \\
        Recovery & \bbar{1.092} \\
        \midrule
        Hispanic & & \\
        \midrule
        LGBTQ & \bbar{1.435} \\
        WCW & \bbar{1.304} \\
        Equality & \bbar{1.282} \\
        IBelieveChristineBlaseyFord & \bbar{1.259} \\
        SESTA & \bbar{1.216} \\
        \midrule
        Asian & & \\
        \midrule
        Brutality & \bbar{4.155}\\
        Friendship & \bbar{4.018} \\
        Humanity & \bbar{3.813} \\
        HumanRight & \bbar{3.470} \\
        Child & \bbar{3.312} \\
        \bottomrule
    \end{tabular}}
    \caption{Top five most distinctive hashtags of each demographic group as measured by pointwise mutual information.}
    \label{tab:hashtags_pmi}
\end{table}

\textbf{Comparing Hashtags Across Race and Ethnicity.} Unlike topics (which were more consistent across gender, but varied more by race and ethnicity), the usage of the most probable hashtags in our dataset vary more by gender and are more consistent across race and ethnicity. However, we observed some notable variations in usage of hashtags across race and ethnicity. For example, as shown in Figure~\ref{fig:hashtags_heatmap}, \#TimesUp is likely to be expressed in tweets authored by black and Hispanic users than among Asian or white users, and the same is true to a lesser extent for \#WhyIDidntReport. 

\begin{quote}
    \textit{``The \#TimesUp and  \#MeToo movement have given women even bigger and louder platforms to use to speak out about the way we are treated !!!''} -- Black Author
\end{quote}

\begin{quote}
    \textit{``I reported it didn't matter I was dismissed \& still get blamed \#WhyIDidntReport''} -- Hispanic Author
\end{quote}

\textbf{Comparing Hashtags Across Gender.} As mentioned above, the top hashtags in our dataset were more vary more across gender. For example, as shown in Figure~\ref{fig:hashtags_heatmap}, the hashtags \#BelieveSurvivors, \#TimesUp, and \#WhyIDidntReport were more probable among tweets authored by females than males, regardless of their race or ethnicity. Meanwhile, hashtags such as \#Trump and \#MAGA were more probable among tweets authored by males than females. Note, at the surface-level, \#Survivor seems to be more likely to be expressed in tweets authored by males, but this is reflective of a difference in semantics. In a random sample of 100 \#Survivor tweets for women, 59 referred to surviving sexual violence and 24 referred to the game show Survivor. In contrast, in a random sample of 100 \#Survivor tweets for men, 64 refer to the game show and 22 refer to domestic and/or sexual abuse.

\textbf{Comparing Hashtags Across Intersectional Identities.}
As shown in Figure~\ref{fig:hashtags_heatmap}, the hashtag \#BelieveSurvivors was more probable among tweets authored by Hispanic females than any other demographic group. The hashtag \#Abuse was more probable among tweets authored by Asian males primarily because the word ``abuse'' was often made into a hashtag in the main text of tweets. The higher probability of this hashtag among black and Asian women is for a similar reason. For example,

\begin{quote}
    \textit{``The \#MeToo campaign was so impactful because it applies to \#women who shared their stories of \#abuse, and so many more.''} -- Asian Male Author
\end{quote}

When observing the most distinctive hashtags of each demographic group (see Table~\ref{tab:hashtags_pmi}), we observe again that tweets authored by black women tended to relate the \#MeToo movement to issues that disproportionately affect people of color (e.g., \#Brutality is often used in tweets discussing police brutality) as well as emotional support, whether directed towards themselves (e.g., \#EmbracingMyself) or others (\#Friendship). For example,

\begin{quote}
    \textit{``Thank you everyone for sharing your stories. Victims need a voice and to be respected! \#MeToo \#DomesticAbuse \#EmbracingMyself''} - Black Female Author 
\end{quote}

Many of the top hashtags that are most distinctive in tweets authored by black males focused on other contemporary issues that notably affect young adults, including \#Bullying, \#Suicide, and \#StopChildAbuse. For example,

\begin{quote}
    \textit{``Crazy how many friends I have with \#MeToo stories from so young. \#StopChildAbuse''} - Black Male Author
\end{quote}

Among tweets authored by white users, we observed more distinction in the usage of hashtags across gender. Tweets authored by white females were more likely to relate \#MeToo to other contemporary issues, such as \#ClimateChange and \#GenderEquality. For example,

\begin{quote}
    \textit{``For real? Adding to the list: \#MuellerTime \#ClimateChangeIsReal \#Resistance \#FBR \#MyBodyMyChoice''} - White Female Author
\end{quote}

Meanwhile, tweets authored by white males were more likely than other intersectional demographics to incorporate religious hashtags (\#Pray, \#Jesus, and \#God). These tweets were not always directly related to Me Too; in fact, in a random subsample of 100 tweets with \#Pray, 38 were related to recovery from addiction. These hashtags were sometimes used to express support for victims, and, more infrequently, \#Pray was employed ironically.

\begin{quote}
    \textit{``This \#WhyIDidntReport is heartbreaking. Will \#pray for victims who seek justice''} - White Male Author
\end{quote}

\begin{quote}
    \textit{``Gonna think and \#Pray for \#Kavanaugh during these challenging times. We won't miss you \#IBelieveChristineBlaseyFord''} - White Male Author
\end{quote}

Among tweets authored by Hispanic men and women, the most distinctive hashtags were all associated with current events and political developments.\footnote{Except \#WCW, which was generally used in long strings of hashtags in tweets that did not consist of any non-hashtag text.} The \#SESTA hashtag refers to the Stop Enabling Sex Traffickers Act, which users in our dataset tended to be against. For example,

\begin{quote}
    \textit{``we can't fix all the problems that led to \#MeToo without treating our sex workers better. \#Resist \#SESTA''} -- Hispanic Female Author
\end{quote}

Many of the other hashtags with high PMI expressed in the tweets of  Hispanic users were employed in the strings of tweets which expressed dissatisfaction about the current political administration. For example,

\begin{quote}
    \textit{``until @realDonaldTrump is voted out I am Puerto Rican, I am a woman, I am \#LGBTQ, I am black, I am \#metoo, I am our environment.''} - Hispanic Female Author
\end{quote}

Finally, tweets authored by Asian females and males were more likely to use hashtags referring to contemporary political issues and discussions, such as \#Brutality and \#VoteThemOut2018.

\section{Discussion}
In our analysis of the first year of the \#Metoo movement on Twitter, we found notable differences in representation and storytelling across demographic groups. Below we contextualize these differences in the larger literature centered on representation, disclosure, and visibility on social media across gender, race, and ethnicity.

In our dataset, we found that the number of men and women actively tweeting in \#MeToo  were nearly equal. However, women authored significantly more tweets, especially during the first week of the movement. In general, white people authored significantly more tweets per-user and tweeted more compared to Asian, black, and Hispanic users. Indeed, Modrek and Chakalov (2019) \cite{Modrek_JMIR_2019} found that disclosures during the first week of \#MeToo were primarily from tweets authored by white and female users. Many scholars have been critical that the \#MeToo exemplified historical exclusion of women of color from feminist movements and our results support that despite the Me Too movement's origin to support women of color, the intersectional identity that tweeted the most was white female users.

The most probable topics expressed in our dataset reflected emotional support (e.g., expressing support for those who have experienced incidence of sexual harassment and violence), discussions centered on politics, and informational support (e.g., sharing news stories that highlight the pervasiveness of sexual harassment and violence). Other studies have highlighted that \#MeToo focused on accusations of sexual misconduct by high-profile figures \cite{pew_howusers} and we observe supporting evidence of this. \#MeToo peaked during the initial week and then again during the accusations of sexual violence perpetrated by Judge Kavanaugh. During the latter, new hashtag activism movements were born including \#WhyIDidntReport and \#BelieveSurvivors. 

Consistent with Hosterman et al. (2018) \cite{hosterman2018twitter}, we also observe differential support strategies across gender -- tweets authored by female users were more likely to express emotional support whereas tweets authored by male users were more likely to express informational support. These differential support strategies were also observed in the operationalization of hashtags. For example, tweets authored by female users were more likely to reflect emotional support and disclosure (\#BelieveSurvivors, \#WhyIDidntReport) whereas tweets authored by men were more likely to reflect contemporary politics using \#MAGA and referring to \#Trump both critically and supportively, similar to to the findings of Vigil-Hayes et al. (2019) \cite{vigilhayes_2019_cscw}. For example, tweets may subversively use \#MAGA to express allyship or to dismiss \#MeToo by perpetuate rape myths, similar to \cite{Gallagher_CSCW_2019}'s findings. We also observed that tweets authored by black users were more likely to express emotional support than other racial/ethnic groups and distinctively use hashtags to reflect this (\#EmbracingMyself, \#Friendship). 

Unlike Modrek and Chakalove (2019) \cite{Modrek_JMIR_2019} who found that tweets authored by white people and women had a higher frequency of disclosure, our dataset reflects that disclosure is a more distinctive topic among black and Hispanic women. The observation that women in general are more likely to disclose personal information on Twitter is consistent with Walton and Rice (2013) \cite{walton2013mediated} and previous works finding that women were approximately twice as likely as men to disclose sexual violence online \cite{Andalibi_2016_disclose,Lokot_2018_sexviolence}. 

Finally, there was distinctive co-opting of \#MeToo among intersectional identities. Tweets authored by white female and male users were both more likely to include contemporary social and political issues. For example, tweets discussing \#ClimateChange were more distinctive among white female users and tweets discussing religion (\#Pray, \#Jesus, \#God) were more distinctive among white male users. In comparison, tweets authored by black female and male users were both more likely to include discussions on race-based discriminatory behaviors and differential treatment in the justice system. This is also reflected in the hashtags (\#Brutality, \#Humanity) that are distinctive among black women users. Similarly, tweets authored by Hispanic female and male users distinctively discussed police brutality. Both of these are likely a factor of historical and ongoing discriminatory practices and dissatisfaction with the police and the justice system because of inequitable treatment \cite{Dowler,biasedlaw1,biasedlaw2,abrams1989gender,ehrenreich1989pluralist,reasonablewoman,intersectionalmetoo2,gomez2020black}. Less is known about how differing races and ethnicities employ Twitter for activism; however, a number of studies have examined how ``Black Twitter'' is used as a digital counterpublic by black people to comment about contemporary events to build collective framing \cite{Graham_2016_blacktwitter,Hill_2018_blacktwitter,clark2014tweet}, and our results exemplify how minority races and ethnicities are co-opting the \#MeToo movement to build on a historical and ongoing counter-narrative. 

\subsection{Implications for Digital Activism Research and Design}
First, we discuss a practical implication of our work to the research community. Our study employed topic modeling to identify emergent topics in our dataset. At the surface-level, it appears that intersectional identities discuss similar topics. For example, there appears to be a convergence of the topics across tweets authored by males and females; however, the hashtags show a divergence. Through a qualitative review of the topics and associated tweets, we find that although users may employ the same language (and hence the appearances of the same topic), the context of their tweet indicates differential framing of the issue at-hand. For example, tweets by female users tended to draw on their personal experiences or share others' personal experiences highlighting the pervasiveness of sexual harassment and violence, whereas tweets by male users tended to discuss sexual harassment and violence in the context of politics, or as a problem that the general public should tackle. This implies that consideration of semantics is important for the framing of digital activism issues and that automated methods that can explore semantics, such as those employed by Chancellor et al. (2018) \cite{Chancellor_2018_norms}, may be more appropriate to disambiguate framing across groups of activists. 

Second, we discuss the implications of our work to the design community to potentially raise marginalized voices participating in digital activism. One of the tenets of digital activism is the desire to democratize narratives and shape collective action. The ability of social media to support a public sphere, defined as ``a constellation of communicative spaces in society that permit the circulation of information, ideas, debates, ideally in an unfettered manner, and also the formation of political will'' \cite{dahlgren2005internet} is contested \cite{colleoni2014echo}. ``Black Twitter'' possesses characteristics of a digital counterpublic (i.e., a place where subordinated social groups can discuss alternative publics \cite{fraser1990rethinking}) that has led to a number of influential hashtag activism movements \cite{Graham_2016_blacktwitter} and enables new forms of activism \cite{Hill_2018_blacktwitter}. However, previous research has shown a `glass ceiling effect' whereby certain demographics receive more visibility. For example, in general, white males have more followers contributing to more visibility and more interaction elevating their tweets in the feed \cite{Messias_2017_demos}. In contrast, Asian and black females have the lowest visibility. Trending topics often suffer from under-representation, persistently from middle-aged black females \cite{chakraborty2017makes}. Ensuring that topics across demographic groups are seen is critical to forming a collective framing for digital activism. 

In our study, we find that sheer volume of tweets authored by white users are over-represented and the topics that they discuss potentially obscure topics important to the framing of \#MeToo, such as racial inequities. This may prohibit the objective of collective framing. Moreover, according to the Social Identity Model of Collective Action, collective efficacy and social identity are important for predicting collective action \cite{van2008toward}. Designers could support those with marginalized voices by developing tools that allow users to leverage their social networks to promote visibility of their tweets. For example, Bisafar (2018) outlines a computing tool that provides informed views of the social capital (i.e., those with influence) of a user's network and allows users to have a birds-eye view of their impact from doing so (i.e., viewing aggregate interactions) \cite{Bisafar_2018_CSCW}. In turn, leveraging social capital to increase influence may help users coalesce a collective identity and framing ultimately translating to focused collective action \cite{benford2000framing,kelly2006protest}.

\subsection{Limitations and Future Work}
Our goal was to explore representation and storytelling across identities in the \#MeToo movement. Our analysis has several limitations, which suggest areas for future work. 

First, our study focuses on approximating large-scale representation of binary gender (female, male) and the most populous races and ethnicities in the U.S. (Asian, black, Hispanic, white). We acknowledge that is an oversimplification of these identities that overlooks non-binary genders, multiracial and multiethnic identities, and less populous racial and ethnic identities. For example, our study does not include non-binary or transgender identities, which is important given that people who hold these identities are subjected to sexual harassment and violence at disproportionately higher rates \cite{trans_stats,trans_stats2}, and given the criticism that the \#MeToo movement included exclusionary language targeting cis-women \cite{trott2020networked,taranaresponse}. 

Second, numerous identities beyond gender, race, and ethnicity play a role in how people elect to use collaborative technologies \cite{kumar_2019_womenshealth} and even participate in digital activism. We do not consider other identities, like sexual orientation or socioeconomic and immigration status, for example. Socioeconomic and immigration statuses play a critical role in one's sense of agency and ability to report incidences of harassment in the workplace \cite{intersectionalmetoo2}. Most \#MeToo analyses are English-centric (including ours) or U.S.-centric \cite{lukose2018decolonizing}, although it is important to consider the varying cultural contexts under which sexual harassment and assault occurs \cite{dartnall2013sexual}. 

Third, topic models only locate wide-scale patterns in themes of social media content; therefore, this method may not recover marginalized conversations that may happen among a subset of the communities our dataset. Additionally, as shown in our results, although some demographic groups discussed the same topic, further qualitative examination showed that despite using similar language, the expressions of the topics were semantically different. We suggest future work should focus on methods that can capture this variation \cite{Chancellor_2018_norms}, especially when comparing discourse across different identities. 

The above limitations suggest that other participatory research methods like participatory design or in-depth interviews may be more apt for studying how these identities participate(d) in digital activism and the barriers that prevent them from safely doing so.

Finally, we limit our analysis to the first year of the \#MeToo movement. Although our study is encapsulates a significantly longer study period than previous works, future research should nonetheless examine how hashtags are revived or co-opted (or not) over even longer time periods.  

\section{Conclusions}
The design of collaborative communication technologies emphasizes the ability of storytelling to aid our understanding of users' perspectives, especially those outside of one's identity \cite{Ogbonnaya_CriticalRaceTheory_CHI_2020}. Some argue that hashtag activism cannot effect real change while others argue that measuring change is difficult to impossible; however, one can infer metrics that indicate growing support \cite{McCafferty_ACM_2011}. With respect to the \#MeToo movement, it is clear in the frequency, content, and participation of postings in this study that the movement was able to grow tremendous social support online. In our study, we find both supportive and critical voices of \#MeToo with most voicing support. However, we also find that under-represented intersectional groups are using \#MeToo to also call attention to other interconnected historical and ongoing movements. Future work in social-media-based movements could observe these under-represented narratives and find ways to ensure that they can contribute democratically to digital activism.

%Utilizing hashtag activism for collective storytelling can spur contentious collective action to achieve a common objective \cite{Hardin_CollectiveAction_1982,Spier_collectiveaction,Tarrow_2011}---in this case, awareness of gendered violence. Using open-access, collaborative communication technologies for hashtag activism, participation and communication can be democratized at all levels, which is necessary for movement building. Intrinsically, this can empower those participating in this process \cite{Servaes_2006}, but this requires participation from all levels and taking a respectful stance when listening to others' experiences \cite{Servaes_2005}. 

\bibliographystyle{ACM-Reference-Format}
\bibliography{paper}

\end{document}